%% file: main.tex
\documentclass[fleqn,10pt]{wlscirep}
\usepackage[utf8]{inputenc}
\usepackage[T1]{fontenc}
\usepackage{array}
\usepackage{soul}
\usepackage{tabularx} 
\UseRawInputEncoding
\usepackage{url}
\usepackage{color}
\usepackage{tabularray}
\usepackage{longtable}
\usepackage{graphicx}
\usepackage{subfig}
\usepackage{rotating}
\usepackage{tikz}
\usepackage{multicol}

\title{A Systematic Investigation of Document Chunking Strategies and Embedding Sensitivity}

\author[1,*]{Muhammad Arslan Shaukat}
\author[1]{Muntasir Adnan}
\author[1]{Carlos C. N. Kuhn}
\affil[1]{Open Source Institute, Faculty of Science and Technology, University of Canberra, Canberra, Australia}
\affil[*]{arslan.shaukat@canberra.edu.au}

\begin{abstract}
We present the first large-scale, cross-domain evaluation of document chunking strategies for dense retrieval, addressing a critical but underexplored aspect of retrieval-augmented systems. 
In our study, 36 segmentation methods spanning fixed-size, semantic, structure-aware, hierarchical, adaptive, and LLM-assisted approaches are benchmarked across six diverse knowledge domains using five different embedding models.
Retrieval performance is assessed using graded relevance scores from a state-of-the-art LLM evaluator, with Normalised DCG@5 as the primary metric (complemented by Hit@5 and MRR). 
Our experiments show that content-aware chunking significantly improves retrieval effectiveness over naive fixed-length splitting. 
The top-performing strategy, Paragraph Group Chunking, achieved the highest overall accuracy (mean nDCG@5 $\approx$ 0.459) and substantially better top-rank hit rates (Precision@1 $\approx$ 24\%, Hit@5 $\approx$ 59\%). 
In contrast, simple fixed-size character chunking as baselines performed poorly (nDCG@5 < 0.244, Precision@1 $\approx$ 2-3\%). 
We observe pronounced domain-specific differences: dynamic token sizing is strongest in biology, physics and health, while paragraph grouping is strongest in legal and maths. 
Larger embedding models yield higher absolute scores but remain sensitive to suboptimal segmentation, indicating that better chunking and large embeddings provide complementary benefits. 
In addition to accuracy gains, we quantify the efficiency trade-offs of advanced chunking. 
Producing more, smaller chunks can increase index size and latency. Consequently, we identify methods (like dynamic chunking) that approach an optimal balance of effectiveness and efficiency. 
These findings establish chunking as a vital lever for improving retrieval performance and reliability. 
Our work offers practical guidance for selecting chunking strategies that maximise knowledge retrieval quality while managing computational costs, informing the design of next-generation retrieval-augmented AI systems.
\end{abstract}
\begin{document}

\flushbottom
\maketitle

\thispagestyle{empty}
\begin{multicols}{2}
\section*{Introduction}

Large language models (LLMs) have reshaped natural language processing through strong generative and reasoning capabilities \cite{RN47, RN46}. However, their knowledge remains bounded by training data and degrades over time, motivating retrieval-augmented generation (RAG) as a means to ground model outputs in external corpora \cite{RN48}. By integrating dense retrieval with generation, RAG systems improve factual accuracy, transparency, and applicability across knowledge-intensive domains, including law \cite{RN49}, finance \cite{RN62}, and medicine \cite{RN51}. Consequently, dense retrieval has become a foundational component of modern knowledge-driven systems \cite{RN5,RN3}.\\

A central yet underexamined component of dense retrieval pipelines is chunking, the segmentation of documents into retrievable units prior to embedding and indexing. Chunking governs how semantic information is represented in vector space, how efficiently similarity search operates, and how faithfully retrieved evidence reflects the source text \cite{RN5,RN53}. Despite its ubiquity, chunking has traditionally been treated as a secondary engineering choice. Recent studies, however, demonstrate that segmentation decisions can substantially affect retrieval effectiveness, latency, memory usage, and index size \cite{RN8,RN55}. Suboptimal chunking can dilute semantic coherence, inflate computational costs, and undermine the benefits of advanced embedding models.\\

Conventional dense retrieval systems rely on fixed-size chunking, segmenting documents into uniform spans of tokens or characters. While simple and reproducible, this approach assumes uniform information density across text. Empirical evidence from long-document retrieval shows that chunk size has a strong and often non-linear effect on performance \cite{RN8,RN58,RN59}. Smaller chunks favour precise matching but risk losing context, whereas larger chunks capture broader semantics at the cost of increased redundancy and query mismatch \cite{RN8}. These effects further interact with embedding architectures: sentence-level encoders and retrieval-tuned contrastive models exhibit distinct optimal segmentation regimes \cite{RN60,RN61}. Such findings indicate that fixed-size chunking is insufficiently flexible for diverse retrieval settings.\\

In response, a growing body of work has proposed semantic and structure-aware chunking strategies. Semantic chunking groups text by meaning similarity to form contextually coherent units \cite{RN9}, while structure-aware methods leverage layout cues such as headings, sections, and metadata to preserve logical boundaries \cite{RN10}. These approaches have shown consistent gains in domain-specific applications, including financial filings \cite{RN62} and legal reasoning tasks \cite{RN49}. More recent advances extend segmentation beyond flat representations. Hierarchical chunking enables multi-resolution retrieval and improved interpretability \cite{RN12}, while late chunking embeds long documents holistically before partitioning, preserving global context \cite{RN3}. Adaptive methods further adjust chunk boundaries based on content complexity or semantic variability \cite{RN65,RN66}. Together, these developments reflect a shift toward content-sensitive segmentation as a key design lever in retrieval systems.\\

Chunking strategies have also been specialised for particular domains, including biomedical literature \cite{RN68}, finance \cite{RN69}, legal documents \cite{RN70}, and code retrieval \cite{RN71}, reinforcing that segmentation is inherently domain-dependent. Parallel efforts have sought to formalise chunking evaluation through protocols and benchmarks such as A New HOPE \cite{RN53} and Hichunk \cite{RN12}, as well as diagnostic tools like RAGTrace \cite{RN40} and optimisation frameworks such as RAGSmith \cite{RN72}. Nonetheless, most existing studies evaluate a limited set of methods, confound chunking with retrieval or embedding choices, or omit efficiency analysis. As a result, key questions regarding cross-domain generalisation, embedding interactions, and operational trade-offs remain unresolved \cite{RN11,RN12}.\\

This paper addresses these gaps through a systematic benchmark evaluating 36 chunking strategies across 6 domains and 5 embedding models (1,080 configurations total) the largest controlled comparison of chunking methods to date. We systematically evaluate thirty-six chunking strategies, including fixed-size, semantic, structure-aware, hierarchical, adaptive, and late chunking, across six domains and five embedding models under a fixed retrieval configuration. Retrieval effectiveness is evaluated using graded relevance judgments produced by large language models, with Normalised Discounted Cumulative Gain (nDCG@5) as the primary ranking sensitive metric, complemented by Hit Rate (Hits@5) and Mean Reciprocal Rank (MRR) to capture retrieval accuracy. In addition, we analyse latency, memory consumption, and index size to characterise efficiency accuracy trade-offs.\\

Our results demonstrate consistent performance hierarchies across domains and embedding models, showing that structurally informed and adaptive chunking strategies frequently outperform fixed-size baselines while exhibiting greater robustness. However, these gains often incur additional computational costs, underscoring the need to balance retrieval quality with operational constraints. Overall, this study reframes chunking as a first-class design dimension in dense retrieval and RAG systems, providing empirical guidance for principled segmentation choices in real-world deployments.

\section*{Material and Methods}

\begin{figure*}[ht!]
\centering
\includegraphics[width = 0.8\linewidth]{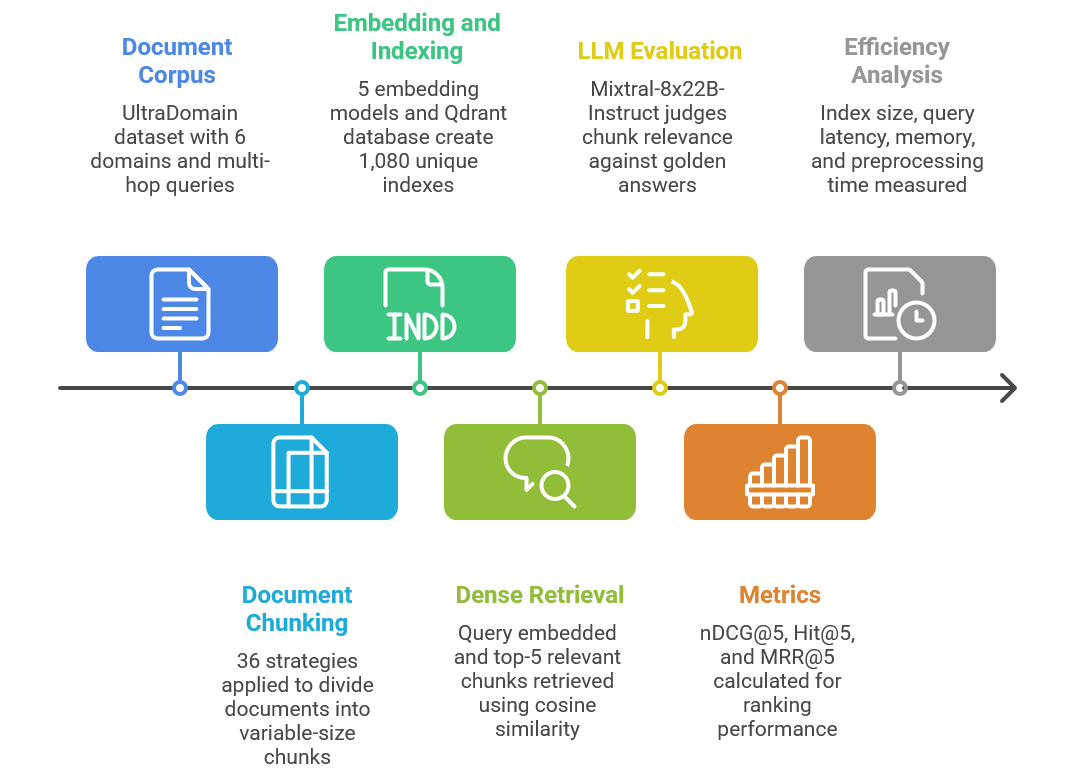}
\caption{End-to-end experimental pipeline for evaluating document chunking strategies in dense retrieval. Documents from six knowledge domains (Biology, Physics, Health, Legal, Maths, Agriculture) in the UltraDomain dataset are segmented using 36 chunking strategies spanning six design categories. The resulting chunks are embedded using five dense embedding models and indexed into separate Qdrant vector stores, yielding 1,080 unique configurations. For each query, the top-5 retrieved chunks are evaluated by a Mixtral-8x22B LLM judge against the golden reference answer using a three-point graded relevance scale. The query is intentionally withheld from the evaluator to prevent lexical bias. Efficiency metrics - including index size, query latency, and memory usage are recorded in parallel across all configurations. No generation component is included; evaluation is retrieval only.}
\label{methodology}
\end{figure*}

This study benchmarks a wide range of document chunking strategies to evaluate their impact on dense retrieval performance across multiple knowledge domains. The full RAG systems combine retrieval with generation, this study isolates the retrieval stage to measure chunking impact without confounding effects from prompt design, decoding, or answer synthesis. Retrieval quality remains the critical bottleneck for RAG success, making this controlled analysis directly relevant. An overview of the complete pipeline is provided in Figure \ref{methodology}.\\

\subsection*{Dataset and Domains}

All experiments are conducted using the UltraDomain dataset \cite{RN1}, a long-context benchmark designed to evaluate retrieval in scenarios where relevant information is distributed across large document collections rather than localised within a single passage. The dataset consists of multi-hop natural-language queries paired with golden reference answers, which specify the information required to satisfy each query.\\

The dataset spans eighteen disciplines; this study focuses on six high-density knowledge domains: Biology, Maths, Physics, Health, Legal, and Agriculture. These domains were selected to capture diverse document structures, ranging from highly formal and hierarchical texts (e.g. legal and mathematical documents) to concept-dense technical narratives (e.g. health and physics). For each domain, the original document contexts provided by UltraDomain are preserved to maintain domain-specific discourse characteristics.\\

Queries drive retrieval of chunks from the vector database, with no golden reference exposure during this step. Golden reference answers are not exposed during chunking, indexing, or retrieval so that it remains answer-agnostic. They are introduced only at the evaluation stage, where they are used to judge whether retrieved chunks support the reference answer.

\subsection*{Chunking Framework Overview}

We evaluate thirty-six document chunking strategies designed to systematically explore the impact of chunk boundary selection on retrieval performance. 
The strategies span a broad design space and vary along three primary axes: 
(i) the mechanism used to determine chunk boundaries (deterministic, structural, semantic, adaptive, or model-driven), 
(ii) the degree of control over chunk size (fixed, overlapping, or variable), and 
(iii) the use of post-processing constraints such as token-length normalisation.\\
 
Each family captures a distinct chunking principle commonly used in information retrieval and neural retrieval pipelines, including fixed-size segmentation, linguistically motivated segmentation, semantic cohesion-based segmentation, adaptive granularity, hierarchical chunking, and LLM-assisted boundary detection. Hybrid strategies arise naturally from combining these principles in multi-stage pipelines.

Table~\ref{table2} provides a complete operational and mathematical specification of each chunking strategy family. 
For each family, the table defines the chunk construction process, all relevant parameters, and the formal criteria used to introduce chunk boundaries.\\

Table~\ref{table1} enumerates the full set of thirty-six implemented strategies, including their abbreviations and concrete parameterisations.
Each listed strategy corresponds to either a direct instance or a parameterised hybrid of the formulations presented in Table~\ref{table2}.
No strategy receives task-specific tuning or special handling.\\

Across all strategies, the downstream retrieval pipeline is held constant. 
Chunks constitute the atomic retrieval unit and are embedded, indexed, and retrieved using identical procedures.

\input{table_1}
\input{table_2}

\subsection*{Embedding Models and Index Construction}

Each chunk produced by a given strategy is embedded using five publicly available models that span high-capacity multilingual transformers, compact sentence encoders, and static retrieval-oriented embeddings. At the high end, we employ BAAI/bge-m3, a 1024-dimensional multilingual embedding model designed for dense, sparse, and multi-vector retrieval and shown to achieve state-of-the-art performance on multilingual and cross-lingual retrieval benchmarks~\cite{RN79}. As a strong but lightweight transformer baseline, we use sentence-transformers/all-MiniLM-L6-v2, a 384-dimensional MiniLM variant distilled for task-agnostic semantic similarity and widely adopted for semantic search and sentence embedding tasks~\cite{RN80,RN81}. We include three static-embedding POTION / Model2Vec models (potion-8M, potion-4M, potion-2M) that provide 256, 128, and 64 dimensional token embeddings respectively, aligning with recent work showing that static word embeddings distilled from Sentence Transformers can yield competitive sentence representations~\cite{RN82}. \\

For each combination of domain, chunking strategy, and embedding model, a separate vector index is constructed using the Qdrant vector database \cite{RN14}. All indices share identical configuration parameters, including distance metric and indexing settings, to ensure fair comparison across conditions and independent analysis of chunking and embedding effects while preserving their interaction.

\subsection*{Dense Retrieval Setup}

Dense retrieval is performed using cosine similarity between query embeddings and indexed chunk embeddings. For each query, the retriever returns the top-5 most similar chunks. This retrieval depth reflects common deployment settings in dense retrieval systems and allows meaningful differentiation between chunking strategies using graded ranking metrics.\\

No sparse retrieval, re-ranking, fusion, or generation-based refinement is applied. The output of the retrieval stage is a ranked list of retrieved chunks for each query, which will serve as input to the evaluation stage along with the golden answer.

\subsection*{Relevance Evaluation and Metrics}

Retrieval quality is evaluated using graded answer-support judgements produced by a large language model acting as an automated assessor. For each query, the evaluator is provided with the retrieved chunks returned by the dense retriever and the corresponding golden reference answer from the dataset. The query itself is intentionally withheld from the evaluator, who is shown only the answer and the retrieved chunk. This setup is designed to assess whether the retrieved chunks support the reference answer, rather than their conventional relevance to query text.\\

The evaluator is asked to assess the extent to which each retrieved chunk contains information that supports or contributes to the golden reference answer. This formulation treats relevance assessment as an answer support attribution task, rather than query-document matching, and is well-suited to the multi-hop and long-context design of the UltraDomain benchmark. The golden reference answer is used exclusively for post-retrieval evaluation and is never exposed to the chunking, indexing, or retrieval stages, ensuring that retrieval performance is evaluated independently of answer knowledge.\\

Prior research has shown that reference based metrics such as BLEU and BertScore, often exhibit weak correlation with human judgements on reasoning-intensive tasks, especially when deep semantic support must be assessed \cite{RN74}. These metrics are primarily sensitive to surface level lexical overlap or static embedding similarity and therefore struggle to distinguish chunks that genuinely support a target answer, as opposed to merely sharing vocabulary or local paraphrases.  In contrast, large language models prompted as evaluators (LLM-as-a-judge) follow natural-language rubrics and directly score properties such as coherence, factual consistency, and answer support, and they consistently achieve much stronger alignment with human preferences across summarisation, dialogue, QA, statistical reasoning, and evaluation tasks \cite{RN78,RN74,RN13,RN75,RN76,RN77}. \\

We employ mistralai/mixtral-8x22b-instruct-v0.1 (via Ollama) as the fixed relevance evaluator. This model is selected based on its top-ranked performance in a recent large-scale LLM judge benchmark measuring agreement with human annotators, where it achieves highly consistent performance relative to humans (Z-score = 1.45 above the mean of all evaluated LLM judges in \cite{RN13}), strong correlation with human relevance scores (Pearson r = 0.879), and high inter-rater reliability (Cohen's k = 0.813). Each retrieved chunk is assigned a graded relevance score on a three-point ordinal scale: \\
\begin{itemize}
    \item \textbf{ Irrelevant, 0:} the chunk does not address or support 
    the reference answer.
    \item \textbf{ Partially relevant, 1:} the chunk is related to the 
    topic but only partially supports the reference answer.
    \item \textbf{ Fully relevant, 2:} the chunk directly and completely 
    supports the reference answer.
\end{itemize}

To ensure deterministic and reproducible outputs, inference was 
conducted with temperature = 0 and top\_p = 0.1. The following 
zero-shot system prompt was provided verbatim to the evaluator:

\begin{verbatim}
You are a strict information retrieval judge.

Reference Answer:
{answer}

Retrieved Chunk:
{chunk_text}

Assign a relevance score:
0 = Not relevant
1 = Partially relevant
2 = Fully relevant

Respond with JSON only:
{
  "score": 0 | 1 | 2,
  "reason": "short explanation"
}
\end{verbatim}

\subsection*{Processing}

For each query $q$ under configuration $(m, d, s)$ comprising embedding model 
$m$, domain $d$, and chunking strategy $s$ we evaluated the top $K = 5$ 
retrieved chunks ordered by rank. We report Normalised Discounted Cumulative Gain at rank 5 (nDCG@5) as the primary 
effectiveness metric, as it jointly captures ranking order and graded relevance. 
Let $g_i \in \{0, 1, 2\}$ denote the gain at rank $i$. The Discounted 
Cumulative Gain is:

\begin{equation*}
    \mathrm{DCG@5}(q, m, d, s) = \sum_{i=1}^{5} \frac{g_i}{\log_2(i + 1)}
\end{equation*}

The discount factor $\log_2(i+1)$ penalises relevant chunks that appear at 
lower ranks, rewarding strategies that surface the most useful content first. 
To normalise across queries with different gain distributions, we compute an 
Ideal DCG (IDCG) by pooling all gains observed for query $q$ across all 
chunking strategies (under fixed $m$ and $d$), sorting them in descending order, 
and applying the same discount:

\begin{equation*}
    \mathrm{IDCG@5}(q, m, d) = \sum_{i=1}^{5} \frac{g_i^{\star}}{\log_2(i + 1)}
\end{equation*}

where $g_1^{\star} \geq g_2^{\star} \geq \cdots \geq g_5^{\star}$ are the top-5 
pooled gains. This defines the ideal ranking relative to what the judge has 
actually assessed, grounding normalisation in observed data. The normalised 
score is then:


\begin{equation*}
    \mathrm{nDCG@5}(q, m, d, s) =
    \begin{cases}
        \tfrac{\mathrm{DCG@5}(q, m, d, s)}{\mathrm{IDCG@5}(q, m, d)} 
        & \text{if } \mathrm{IDCG@5} > 0 \\[6pt]
        0 & \text{otherwise}
    \end{cases}
\end{equation*}

Binary metrics, including Hit@5 and MRR@5, are reported for complementary analysis and to facilitate comparison with prior retrieval benchmarks. We adopt a strict relevance criterion for binary metrics: only chunks 
judged \textit{fully relevant} (score $= 2$) are counted as relevant. Partially 
relevant chunks (score $= 1$) are treated as non-relevant for Hit@5 and MRR, 
avoiding inflation of success rates by near-miss retrievals.\\

Hit@5 is a binary indicator of whether at least one fully relevant chunk appears 
in the top 5 results:

\begin{equation*}
    \mathrm{Hit@5}(q, m, d, s) = 
    \mathbf{1}\left[\,\exists\; i \leq 5 : g_i = 2\,\right]
\end{equation*}

Averaged across queries, this yields the Hit Rate, which captures the proportion 
of queries for which retrieval was at least fully successful in the strictest 
sense.\\

Mean Reciprocal Rank measures how early a fully relevant chunk first appears in 
the ranked list. Let $r^{\star}$ denote the rank of the first chunk with $g = 2$:

\begin{equation*}
    \mathrm{MRR@5}(q, m, d, s) =
    \begin{cases}
        \dfrac{1}{r^{\star}} & \text{if } \mathrm{Hit@5} = 1 \\[6pt]
        0 & \text{otherwise}
    \end{cases}
\end{equation*}

A strategy that consistently places a fully relevant chunk at rank 1 achieves 
MRR $= 1.0$; placement at rank 3 yields $0.33$. Precision@1 is defined as 1 if the rank~1 retrieved chunk receives a relevance score of 2 (fully relevant), and 0 otherwise.

\subsection*{Efficiency and Resource Metrics}

In addition to retrieval effectiveness, we measure efficiency at two stages of the pipeline. Prepossessing metrics include chunking-phase runtime and peak memory usage during segmentation/index preparation. Retrieval metics include total number of chunks, resulting index size, and query-time latency (median and 95th percentile) which allows us to distinguish one-time indexing cost from recurring query-time cost. 

Latency is measured end-to-end for the retrieval stage, excluding offline preprocessing and indexing. These metrics enable explicit analysis of trade-offs between retrieval quality and operational cost.

\subsection*{Experimental Protocol and Reproducibility}

All experiments follow a fixed evaluation protocol. Each configuration is executed once, reflecting realistic deployment behaviour for retrieval systems. While LLM-based chunking and relevance evaluation introduce stochasticity, all model versions, prompts, and parameters are held constant across configurations. Intermediate artefacts, including chunk outputs, retrieval rankings, and evaluation scores, are stored to ensure traceability and post-hoc analysis.

\section*{Results and Discussion} 

This section analyses the impact of chunking strategies on retrieval effectiveness, robustness, and efficiency under a fixed dense retrieval configuration. Prior studies have suggested that segmentation choices play a substantial role in retrieval performance, yet existing evidence is often fragmented across domains or confounded with changes in retrieval architectures and embedding models \cite{RN11,RN12}. By evaluating a diverse set of chunking strategies across six domains and five embedding models in a controlled setting, our results enable a direct comparison of segmentation approaches and provide a unified view of how chunking strategy performance under a fixed dense retrieval pipeline, across six domains and five embedding models as shown in Figure \ref{supp_figure2}. In doing so, the analysis corroborates earlier observations regarding the limitations of fixed-size chunking \cite{RN8,RN58} and clarifies the conditions under which semantic and structure-aware strategies yield consistent gains.

\subsection*{Overall Chunking Effectiveness}

\begin{figure*}[ht!]
\centering
\includegraphics[width = \linewidth]{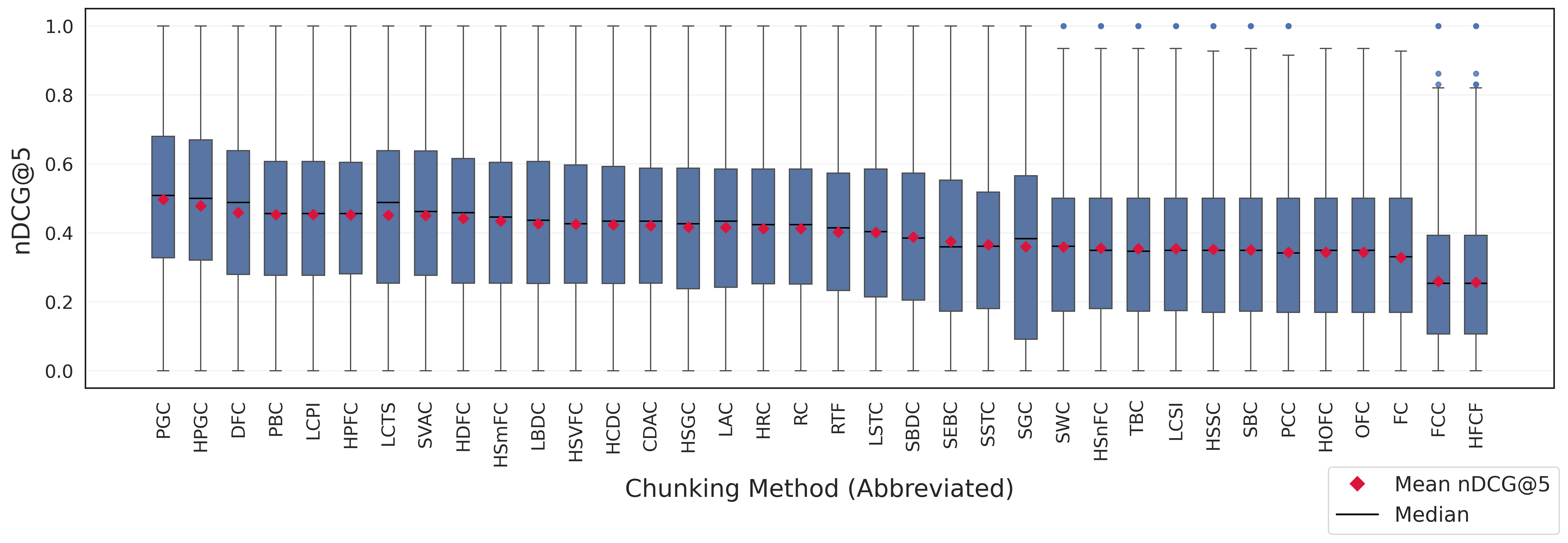}
\caption{Distribution of nDCG@5 for all chunking strategies across embedding models and domains, ordered by descending mean nDCG@5. Red diamonds denote mean values and black lines denote medians. Higher-ranked methods such as PGC, HPGC, and DFC show stronger overall retrieval effectiveness, while FC, FCC, and HFCF exhibit weaker performance. Differences in box and whisker widths highlight stability variations acroos queries. 
The chunking method abbreviations on the x-axis correspond to those defined in Table~\ref{table1}}
\label{query_variance}
\end{figure*}

Across all evaluated domains and embedding models, Paragraph Group Chunking (PGC) achieved the highest mean nDCG@5 ($\approx 0.459$; Figure~\ref{query_variance}), outperforming the next-best Dynamic Token Size Chunking (DFC; 0.441). PGC also led on strict top-rank success (Precision@1 = $24\%$, where Precision@1 = 1 if rank~1 chunk scores fully relevant) and Hit@5 ($59\%$). Fixed character chunking (FCC), a naive baseline, trailed substantially with nDCG@5 = 0.244 and Precision@1 $\approx 2-3\%$. These gaps highlight PGC's ability to surface fully relevant content both early and reliably.\\

By contrast, the second-tier methods Dynamic Token Size Chunking (nDCG@5 = 0.441) and LLM Boundary Detection Chunking (nDCG@5 = 0.415) attain strong but slightly lower overall effectiveness.\\
%

Nearly all advanced chunking strategies substantially outperform the simplest baselines. For example, fixed-length character chunking (FCC) yields one of the weakest outcomes, with mean nDCG@5 = 0.244 and only $2-3\%$ of queries retrieving a relevant document at rank~1. In contrast, the strongest methods achieve 10.4 times higher Precision@1 and markedly higher recall. A query-level comparison further illustrates that modern chunking strategies outperform fixed-character segmentation on a large majority of queries.\\

These findings are consistent with prior work reporting poor retrieval effectiveness for fixed-size segmentation due to excessive context fragmentation \cite{RN8,RN58}. At the same time, they extend earlier domain-specific evidence \cite{RN10,RN49} by demonstrating that both preserving document-native structural coherence (as with PGC) and adapting dynamically to content density (as with DFC) yield strong and stable improvements across multiple domains and embedding models within a unified evaluation framework. Notably, the effectiveness of Paragraph Group Chunking suggests that maintaining coherent logical units can be more beneficial than introducing additional semantic partitioning noise, particularly when embedding models already encode substantial contextual information.\\

Examining methods by their design philosophy further highlights these differences. Strategies that preserve semantic or structural units, such as paragraph-based grouping and dynamic or adaptive chunking, exhibit substantially  higher effectiveness than approaches relying on fixed-size slices. The fixed-length chunking family tends to fragment coherent content or group unrelated text, which plausibly explains its consistently poor performance relative to more content-aware approaches.

\subsection*{Precision vs. Recall Trade-offs}

Although the best-performing chunking strategies tend to improve both precision and recall metrics, certain methods emphasise one at the expense of the other. Highly fine-grained semantic chunking variants that generate many small segments often ensure that some relevant content is retrieved within the top results (boosting Hit@5) but may scatter relevant information across multiple chunks such that none receives the top rank (reducing Precision@1). This pattern is evident for Semantic Similarity Threshold chunking in the legal domain, where the method attains a relatively high Hit@5 of approximately $54\%$ but a lower Precision@1 of around $17\%$.\\

By contrast, Paragraph Group Chunking in the legal domain not only retrieves relevant content frequently (Hit@5 $\approx 78\%$) but also places the most relevant result at rank~1 substantially more often (Precision@1 $\approx 35\%$). Overall, the strongest strategies in our study, Paragraph Group Chunking, LLM-based segmentation, and dynamic approaches, achieve a more favourable balance by producing chunks that are neither overly fragmentary nor excessively coarse. This balance allows embedding models to both identify relevant information early and retain sufficient contextual coverage within the top ranks. Our results further indicate that structure-preserving chunking can mitigate this trade-off more effectively than purely semantic or aggressively fine-grained approaches, maintaining both strong early ranking performance and high coverage of relevant content.

\subsection*{Domain-Specific Performance Trends}

The pronounced domain dependence observed here helps contextualise previously fragmented findings in the literature, where different chunking strategies were reported as optimal for specific domains such as biomedical \cite{RN68} or legal text \cite{RN49} but were not evaluated under comparable retrieval conditions.\\

\begin{figure*}[ht!]
\centering
\includegraphics[width=\linewidth]{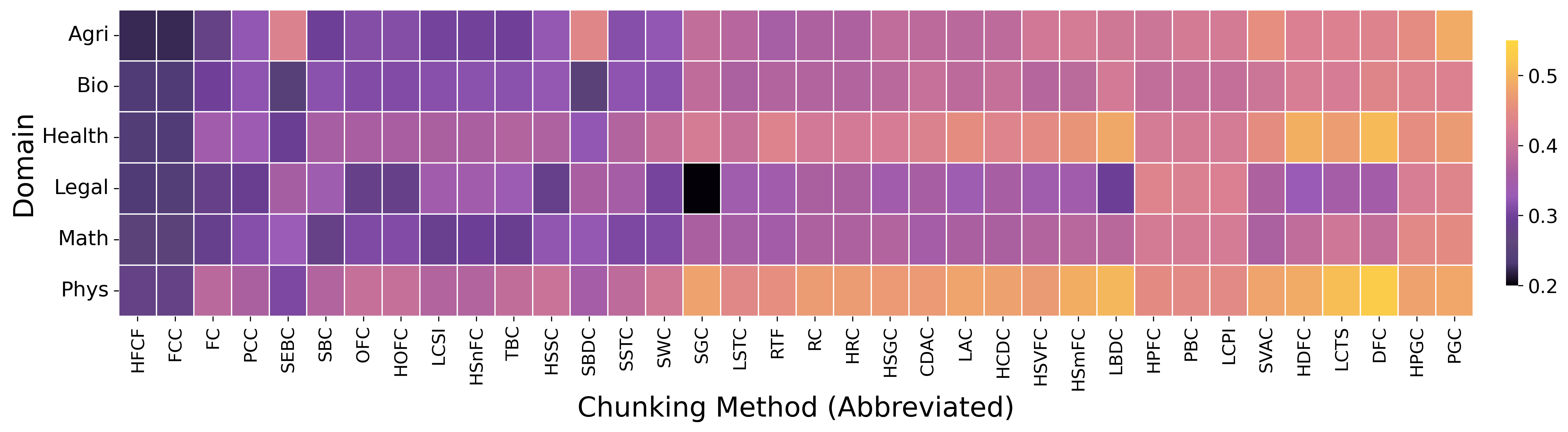}
\caption{Mean nDCG@5 scores for Domain-specific retrieval performance across Agriculture, Biology, Health, Legal, Maths and Physics. Dynamic Token Size Chunking (DFC) achieves the highest single score in Health and Physics while paragraph-aware and late-stage strategies, particularly Paragraph Group Chunking (PGC), Hybrid Paragraph Group Fixed Token Chunking (HPGC), and Late Chunking Token Spans (LCTS), consistently rank among the top three methods across all domains. 
The chunking method abbreviations on the x-axis correspond to those defined in Table~\ref{table1}.}
\label{domain_heatmap}
\end{figure*}

Figure~\ref{domain_heatmap} compares chunking strategies across domains. In biology and physics, Dynamic Token Size Chunking yields the highest average performance, achieving high MRR and Precision@1 across both smaller and larger embedding models. These gains suggest that dynamic adaption to content density align well with semantically meaningful sections in scientific texts, enabling more complete contextual units to be retrieved.\\

In health-related documents, Dynamic Token Size Chunking performs best across all models, likely reflecting the wide variation in document length and structure in this domain. By adapting chunk size to content, this method avoids inappropriate splits of clinically relevant passages, leading to consistently high retrieval effectiveness.\\

For legal and mathematical corpora, Paragraph Group Chunking dominates across all embedding models. Legal documents frequently present arguments spanning multiple paragraphs, while mathematical texts often distribute definitions, theorems, and proofs across contiguous sections. Preserving these multi-paragraph structures enables the retrieval of complete logical units, substantially improving accuracy relative to more fine-grained segmentation strategies.\\

Agricultural texts exhibit a heterogeneous pattern with no single dominant method. Paragraph-aware strategies (PGC, HPGC) and late chunking (LCTS) rank among the top three across embedding models, though semantic methods also perform competitively depending on model capacity. Larger models benefit from embedding-based semantic clustering, while smaller models perform better with explicit semantic boundary detection. These results suggest that agricultural documents, which vary widely in structure and topic, benefit from chunking strategies that enforce topical cohesion, with model capacity influencing which semantic cues are most effectively exploited.\\

Overall, these domain-specific trends reinforce the conclusion that no single chunking strategy is universally optimal; instead, segmentation effectiveness depends on the structural and semantic characteristics of the target domain.

\subsection*{Influence of Embedding Model Size}

\begin{figure*}[ht!]
\centering
\includegraphics[width=\linewidth]{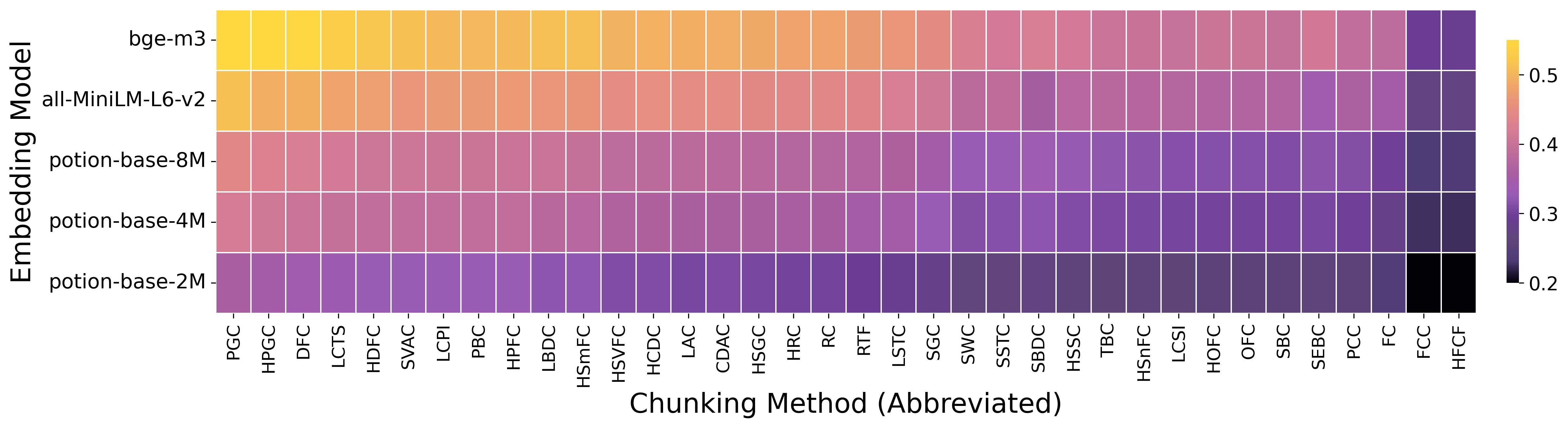}
\caption{Heatmap of mean nDCG@5 across five embedding models and thirty-six chunking strategies. The results show that bge-m3 (0.456) delivers the strongest overall performance, with all-MiniLM-L6-v2 (0.416) as the clear second best. The potion-base variants trail behind, with scores generally tapering off across later chunking strategies.
The chunking method abbreviations on the x-axis correspond to those defined in Table~\ref{table1}.}
\label{embeddingmodel_x_chunkingmethod_heatmap}
\end{figure*}

As expected, larger embedding models achieve higher absolute retrieval effectiveness across all chunking strategies. However, improved segmentation yields gains regardless of model size, and the relative ordering of chunking strategies remains largely stable as model capacity increases (Figure~\ref{embeddingmodel_x_chunkingmethod_heatmap}).\\


For example, the same paragraph-based and adaptive strategies that perform well with all-MiniLM-L6-v2 also tend to perform well with BGE-M3, albeit at a higher absolute nDCG@5. With an optimal chunking strategy, the strongest model reaches very high effectiveness in some domains (e.g., nDCG@5 $\approx 0.65$-$0.70$ in physics), while even the smallest model benefits from the same domain-preferred segmentation approach, though at lower absolute performance levels.\\

These trends align with prior observations that increased embedding capacity can reduce but does not eliminate sensitivity to chunking granularity \cite{RN61}. Even for high-capacity encoders, suboptimal segmentation continues to impose a ceiling on retrieval effectiveness, indicating that embedding quality and chunking strategy play complementary roles rather than serving as substitutes.

\subsection*{Robustness and Query-Level Variability}

Beyond aggregate metrics, we examine the consistency of each chunking strategy across individual queries. Some methods exhibit high variance, performing exceptionally well on certain queries but failing on others. Strategies that aggressively fragment content tend to show greater query-by-query variability. In the physics domain, Dynamic Token and LLM-based chunking achieve high average scores but also exhibit a higher incidence of zero-hit queries, where no relevant result is retrieved, likely because the smaller chunks lacked sufficient global context to meaningfully match the dense query vector, causing them to fall completely out of the top-5 ranking.\\

In contrast, structurally coherent approaches such as Paragraph Group Chunking demonstrate more stable behaviour, with fewer catastrophic failures on difficult queries. A quantitative analysis of query-level variance (Figure~\ref{query_variance}) supports this observation, showing that the strongest methods combine high mean effectiveness with relatively low variability.\\


While most prior evaluations of chunking strategies focus on aggregate retrieval metrics, this analysis highlights meaningful differences in robustness across queries, suggesting that segmentation choices also influence the reliability of retrieval outcomes for diverse information needs.

\subsection*{Effectiveness--Efficiency Trade-offs}

The choice of chunking strategy has significant implications for index size, storage cost, and query latency. Different methods produce widely varying numbers of chunks per document, directly affecting indexing overhead and retrieval efficiency. 
\begin{figure*}[ht!]
\centering
\includegraphics[width = \linewidth]{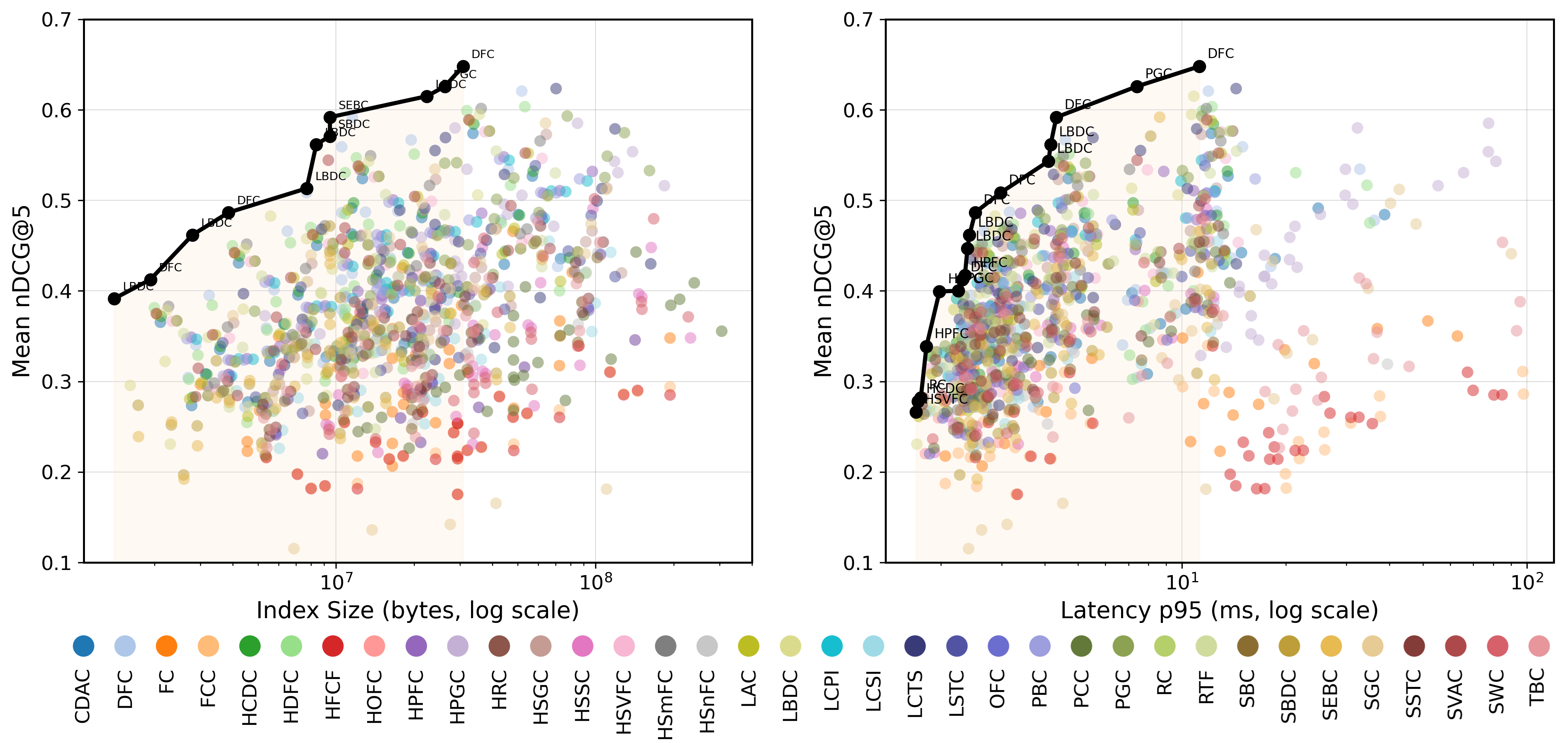}
\caption{Effectiveness-efficiency trade-off plots. Left: mean nDCG@5 against index size, highlighting how each strategy scales in terms of index growth. Right: mean nDCG@5 against query latency, illustrating the trade-off between retrieval accuracy and response time for each chunking approach.
The chunking method abbreviations on the x-axis correspond to those defined in Table~\ref{table1}.}
\label{pareto}
\end{figure*}

While methods that generate many small chunks can achieve high recall, they often incur substantial indexing and latency costs. Conversely, grouping content into very large chunks reduces index size but risks diluting relevance signals. The average chunk length alone is therefore an incomplete predictor of performance; extremely small or large chunks tend to perform poorly, whereas moderate chunk sizes often produced by adaptive or structure-aware methods strike a more favourable balance.\\

To identify strategies that jointly optimise effectiveness and efficiency, we examine the Pareto frontier of mean nDCG@5 versus index size (Figure~\ref{pareto} Left). Only a small subset of methods lies on this frontier, indicating that many chunking strategies are dominated by alternatives offering equal or better effectiveness at lower cost. Dynamic Token Chunking, in particular, offers a strong trade-off, achieving near-optimal retrieval effectiveness while producing substantially fewer chunks than fine-grained approaches. In contrast, the Fixed Character baseline lies far from the frontier, inflating index size while delivering poor effectiveness. \\

A similar pattern emerges when considering query latency (Figure~\ref{pareto} Right), where leaner indexing strategies generally enable faster retrieval. However, index size and query latency only capture system performance during the final retrieval phase; the initial segmentation of documents also incurs heavily skewed computational overheads. As detailed in Table \ref{table3}, the memory footprint (RAM) and processing time required during the chunking phase vary wildly depending on the strategy's design philosophy. LLM assisted boundary detection (LBDC) and segment-then-chunk (LSTC) methods introduce massive preprocessing costs. Similarly, deep semantic processing methods, such as Topic Based Chunking (TBC), drastically inflate computation time, taking over 133 seconds to execute compared to the 6 seconds required by PGC. Although this extreme LLM and semantic overhead is typically amortised as a one time indexing cost and may be entirely acceptable for static corpora where retrieval gains are substantial it presents a severe bottleneck for real-time or frequently updated databases. Jointly analysing these preprocessing constraints alongside retrieval quality addresses a critical gap in prior chunking studies. Paragraph Group Chunking (PGC) lies near the Pareto frontier, delivering top effectiveness with relatively low chunking time, RAM usage, and query latency. Adaptive methods like Dynamic Token Size Chunking (DFC) offer a competitive alternative for domains favoring variable granularity, though they incur modestly higher preprocessing costs.

\input{table_3}

Taken together, these findings demonstrate that chunking is a central design factor in dense retrieval systems, influencing not only average effectiveness but also robustness across queries and efficiency at scale. While Paragraph Group Chunking achieves the strongest overall performance (nDCG@5 = 0.459), Dynamic Token Size Chunking proves advantageous in Biology (0.534), Health (0.621), and Physics (0.648). LLM-informed segmentation, dynamic resizing, and semantic clustering, prove advantageous in specific domains, reinforcing prior observations that optimal segmentation is inherently context-dependent \cite{RN49,RN62,RN68}. The consistency of these trends across embedding models further supports the view that effective chunking amplifies, rather than replaces, the benefits of stronger embeddings \cite{RN61}. By situating these results within a unified evaluation framework, this study clarifies how prior domain-specific insights generalise and provides practical guidance for selecting chunking strategies that balance retrieval quality with operational constraints in real-world retrieval-augmented systems.

\subsection*{Metric Validation and Correlation Structure}

\begin{figure*}[htbp!]
\centering
\includegraphics[width = 0.5\linewidth, height=8cm]{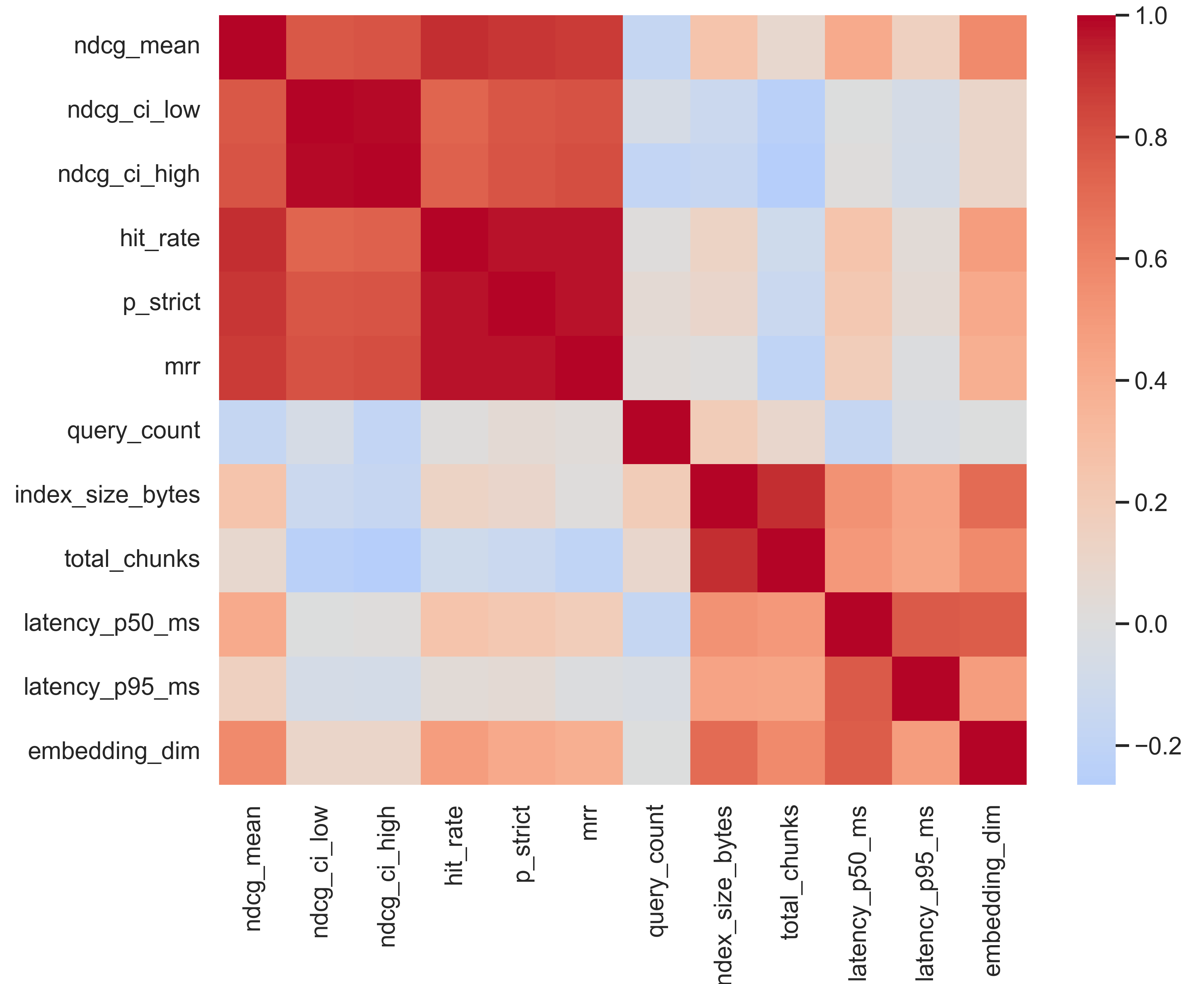}
\caption{Pearson correlation matrix across all evaluation metrics. The nDCG@5 family shows strong internal consistency and high correlation with MRR@5 (r=0.92) and Precision@5 (r=0.88), validating it as the primary ranking metric. 
Embedding dimensionality shows minimal correlation with performance metrics (|r| < 0.2), confirming that chunking strategy selection dominates model architecture as a performance driver. Efficiency metrics (index size, latency) occupy the lower-right quadrant and show expected negative trade-off relationships.}
\label{supp_figure2}
\end{figure*}

Figure \ref{supp_figure2} presents a comprehensive Pearson correlation matrix visualising relationships across key evaluation metrics from the chunking strategy benchmark. It serves three primary purposes: validating metric reliability, identifying complementary measures, and quantifying performance-efficiency trade-offs.\\

The nDCG@5 family (mean, low/high confidence intervals) occupies the upper-left quadrant, demonstrating exceptional internal consistency. nDCG\_mean exhibits the strongest correlation with MRR@5 (r=0.92), confirming that strategies producing high-quality early rankings also achieve superior reciprocal rank performance. Strict Precision@5 (p\_strict) follows closely (r=0.88), validating that top-ranked relevance directly translates to precision. The confidence interval metrics (ndcg\_low, ndcg\_high) show moderate mutual correlation (r$\approx$0.6), indicating stable variance patterns across strategies. This tight clustering confirms nDCG@5 as a robust primary metric suitable for primary reporting.\\

Binary success metrics form a secondary cluster. Hit@5 correlates strongly with nDCG\_mean (r=0.85) and MRR@5 (r=0.82), establishing that strategies succeeding on at least one fully relevant chunk in top-5 also excel in ranking quality. MRR@5's central position underscores its role as a ranking-focused complement to nDCG's graded assessment.\\

Efficiency metrics reveal expected engineering trade-offs in the lower-right quadrant. Index size exhibits moderate positive correlation with both latency metrics (latency\_p95: r$\approx$-0.4, latency\_p90: r$\approx$-0.35), confirming larger indexes slow retrieval as expected. Total chunk count shows similar positive latency relationships, validating that fine-grained chunking increases computational overhead. 
Embedding dimensionality displays minimal correlation with performance metrics (|r|<0.2), suggesting chunking strategy selection dominates model architecture choice.

\subsection*{Limitations and Future Work}

A few limitations should be noted. 
First, chunking strategies inherently affect index size and chunk count, which may influence retrieval behaviour beyond semantic quality alone. Rather than normalising these effects away, we explicitly measure and analyse them as part of the system-level trade-offs reported in the Results and Discussion section. Second, relevance judgements are produced by an LLM rather than human annotators. Although the selected evaluator demonstrates strong agreement with humans, automated evaluation may still introduce bias or noise. Third, results are reported on a fixed subset of domains from UltraDomain; while these domains are diverse, findings may not generalise to all retrieval settings. Future work could incorporate ensembles of judges, calibration procedures, or targeted human spot-checks to further strengthen reliability in large-scale retrieval benchmarks.

\section*{Conclusion}
This study demonstrates that document chunking is a pivotal design factor in dense retrieval systems. In our cross-domain evaluation of 36 chunking strategies across six knowledge domains and five embedding models, we find that segmentation choices significantly affect retrieval accuracy, robustness, and efficiency. Content-aware approaches that preserve natural semantic or structural units consistently outperform fixed-size baselines, confirming that naive uniform segmentation can severely degrade performance. For example, the best method, Paragraph Group Chunking, achieved the highest overall retrieval quality (mean nDCG@5 $\approx$ 0.459) along with substantially higher top-rank precision ($\approx$ 24\% Precision@1) and coverage ($\approx$ 59\% Hit@5) than other strategies. In contrast, a simple fixed-length character chunking baseline produced a mean nDCG@5 below 0.244 with only $\approx$ 2-3\% of queries retrieving a relevant document at rank 1. These results underscore the risk of arbitrarily fragmenting context and highlight the value of segmentation that aligns with the text's logical units.\\

No single chunking strategy is universally optimal; instead, effective chunking is context-dependent, varying with domain and content characteristics. Strategies such as LLM-informed boundary detection, dynamic token sizing, and paragraph grouping each excelled in different domains, reflecting the diverse structure of scientific, clinical, legal, and technical texts. Nonetheless, the performance hierarchy remains fairly consistent across embedding model sizes. Even the largest models benefit from improved chunking, and suboptimal segmentation imposes a performance ceiling on all models. This indicates that chunking and embedding capacity offer complementary benefits: stronger embeddings amplify the gains from good chunking rather than replacing the need for it. This means investing in a larger embedding model without revisiting chunking strategy is an incomplete optimisation.\\

Our results also highlight practical trade-offs between retrieval effectiveness and efficiency. Methods producing extremely fine-grained chunks can boost recall but at the cost of exploding index size and query latency, whereas overly coarse chunks risk missing relevant information. We identified adaptive strategies, such as dynamic token chunking, that lie near the Pareto-optimal frontier, delivering high accuracy with fewer chunks, unlike naive baselines, which inflate resource usage for inferior results. These insights underscore the need to balance quality and operational constraints when selecting a chunking policy for real-world applications.\\

Taken together, this work reframes chunking from a low-level implementation detail to a first-class consideration in retrieval-augmented systems. By providing a unified, large-scale evaluation across domains, we bridge previously fragmented findings and offer empirical guidance for choosing segmentation strategies. Careful, content-aware chunking can substantially improve retrieval success and consistency and as retrieval-augmented systems scale to larger corpora and more demanding applications, the choice of how to divide knowledge will remain as consequential as the choice of how to represent it.

\section*{Data Availability}
All benchmark datasets used in this study are publicly available.
The code and analysis scripts will be made publicly available upon publication at \url{https://github.com/TheOpenSI/chunkbench}.

\bibliography{sample}

\section*{Acknowledgments}
Portions of the manuscript text and implementation were developed with the assistance of an AI-based language model. The authors designed the methodology, conducted all experiments, and take full responsibility for the results and conclusions presented in this work.

\section*{Author contributions}
M.A.S, C.C.N.K and M.A collaborated on exploring the idea, writing, designing the experiments and reviewing the manuscript. All authors reviewed the manuscript.

\section*{Funding}
This work is funded under the agreement with the ACT Government, Future Jobs Fund - Open Source Institute (OpenSI) - R01553 and NetApp Technology Alliance Agreement with OpenSI - R01657.
Additionally, this research was supported by the Australian Government through the Department of Education's National Industry PhD Program (project 36337). 
The views expressed herein are those of the authors and are not necessarily those of the Australian Government or the Department of Education.

\section*{Competing interests}
All authors declare that there is no conflict of interest and no financial contributions have been made towards this work, which could have affected its results.

\newpage

\appendix

\end{multicols}

\end{document}

%% file: table_1.tex
\begin{table*}[htbp]
\centering
\small
\setlength{\tabcolsep}{6pt}
\caption{Strategy configuration matrix showing chunking category, strategy name, shorthand alias used throughout the paper, and the corresponding operational configuration derived from the experimental setup.}
\begin{tabular}{p{4.1cm} p{5.4cm} p{1.6cm} p{6.7cm}}
\toprule
\textbf{Category} & \textbf{Strategy Name} & \textbf{Abbreviation} & \textbf{Configuration Set} \\
\midrule

\textbf{Deterministic / Rule-Based}
& Fixed Character Chunking & FCC & char\_size=100, overlap=10 \\
& Fixed Token Chunking & FC & token\_size=50, overlap=0 \\
& Overlapping Token Chunking & OFC & token\_size=50, overlap=10 \\
& Sliding Window Token Chunking & SWC & window=50, step=25 \\
& Length Aware Chunking & LAC & target=500, tolerance=100 \\
& Sentence Based Chunking & SBC & sentence boundaries \\
& Sentence Group Chunking & SGC & sentences=3, overlap=1 \\
& Paragraph Based Chunking & PBC & paragraph boundaries \\
& Paragraph Group Chunking & PGC & paragraphs=2, overlap=1 \\
\midrule

\textbf{Recursive / Hierarchical}
& Recursive Chunking & RC & chunk\_size=500, overlap=50 \\
& Recursive Token Fallback Chunking & RTF & token\_size=100, overlap=10 \\
& Parent Child Chunking & PCC & parent=500, child=100 \\
\midrule

\textbf{Semantic / Topic-Aware}
& Semantic Embedding Based Chunking & SEBC & embedding similarity grouping \\
& Semantic Similarity Threshold Chunking & SSTC & similarity threshold=0.6 \\
& Topic Based Chunking & TBC & topic distance threshold=0.4 \\
& Semantic Boundary Detection & SBDC & semantic boundary detection \\
\midrule

\textbf{Adaptive / Dynamic}
& Dynamic Token Size Chunking & DFC & min=50, max=200 tokens \\
& Content Density Adaptive Chunking & CDAC & base\_size=1000 \\
& Semantic Variance Adaptive Chunking & SVAC & sensitivity=0.2 \\
\midrule

\textbf{Late Chunking / Index-First}
& Late Chunking Sentence Indexing & LCSI & sentence-level indexing \\
& Late Chunking Paragraph Indexing & LCPI & paragraph-level indexing \\
& Late Chunking Token Spans & LCTS & span=128, step=64 \\
\midrule

\textbf{LLM-Driven}
& LLM Boundary Detection Chunking & LBDC & LLM-based boundary inference \\
& LLM Segment Then Chunk & LSTC & LLM segmentation + size constraint \\
\midrule

\textbf{Semantic-First Hybrids}
& Hybrid Semantic Fixed Token Chunking & HSmFC & semantic $\rightarrow$ token(200,20) \\
& Hybrid Semantic Sliding Window Chunking & HSSC & semantic $\rightarrow$ window(50,25) \\
& Hybrid Semantic Variance Fixed Token Chunking & HSVFC & variance $\rightarrow$ token(200,20) \\
\midrule

\textbf{Structural-First Hybrids}
& Hybrid Sentence Fixed Token Chunking & HSnFC & sentence $\rightarrow$  token(200,20) \\
& Hybrid Sentence Group Fixed Token Chunking & HSGC & sent\_group $\rightarrow$ token(200,20) \\
& Hybrid Paragraph Fixed Token Chunking & HPFC & paragraph $\rightarrow$ token(200,20) \\
& Hybrid Paragraph Group Fixed Token Chunking & HPGC & para\_group $\rightarrow$ token(200,20) \\
& Hybrid Recursive Fixed Token Chunking & HRC & recursive $\rightarrow$ token(200,20) \\
& Hybrid Fixed Char Fixed Token Chunking & HFCF & char(100,10) $\rightarrow$ token(200,20) \\
& Hybrid Overlapping Fixed Token Chunking & HOFC & overlap(50,10) $\rightarrow$ token(200,20) \\
& Hybrid Dynamic Fixed Token Chunking & HDFC & dynamic(50-200) $\rightarrow$ token(200,20) \\
& Hybrid Content Density Fixed Token Chunking & HCDC & density $\rightarrow$ token(200,20) \\
\bottomrule
\end{tabular}
\noindent\textbf{Configuration legend.}
Numeric pairs are reported as \texttt{(size, overlap)} in tokens unless otherwise stated.
Ranges (e.g., \texttt{50-200}) denote adaptive minimum and maximum token limits.
\texttt{window(step)} denotes sliding window size and stride.
Arrows (\(\rightarrow\)) indicate sequential hybrid execution, where the primary chunking strategy is applied first and the output is subsequently normalised by the secondary strategy.
Structural boundaries (sentence/paragraph) imply no fixed size unless combined with token normalisation.
Similarity thresholds and sensitivities are cosine-based unless otherwise noted.

\label{table1}
\end{table*}

%% file: table_2.tex
\begin{table*}[htbp]
\centering
\small
\setlength{\tabcolsep}{5pt}
\caption{Detailed specification of chunking strategies evaluated in this study. Each row provides a complete operational and mathematical definition; all implemented strategies correspond to parameterised instances or hybrids of these formulations.}
\renewcommand{\arraystretch}{1.2}
\begin{tabular}{p{3cm} p{6.5cm} p{6.5cm}}
\toprule
\textbf{Strategy} &
\textbf{Operational Definition} &
\textbf{Formal Specification and Parameters} \\
\midrule

Fixed Token Chunking (FC) &
The document is tokenised and split into contiguous, non-overlapping chunks of fixed length. Each chunk is treated as an independent retrieval unit. &
Let $T=[t_1,\dots,t_N]$ be the token sequence of document $D$.  
The $i$-th chunk is defined as  
$C_i = T[(i-1)K : (i-1)K + K]$,  
where $K$ is the fixed number of tokens per chunk. \\

Fixed Character Chunking (FCC) &
The raw document text is divided into contiguous character spans of equal length, independent of linguistic structure. &
Let $D$ be a character sequence.  
$C_i = D[(i-1)L : (i-1)L + L]$,  
where $L$ is the fixed number of characters per chunk. \\

Sliding Window Token Chunking (SWC / OFC) &
Overlapping token windows are generated to preserve contextual continuity across chunk boundaries. &
$C_i = T[(i-1)S : (i-1)S + W]$,  
where $W$ is the window size (tokens) and $S$ is the stride.  
Overlap occurs when $S < W$. \\

Sentence-Based and Sentence-Group Chunking (SBC / SGC) &
Chunks are formed by grouping one or more consecutive sentences, preserving sentence boundaries. &
Let $\{s_1,\dots,s_n\}$ denote sentences in $D$.  
$C_i = \{s_{(i-1)(G-O)+1},\dots,s_{(i-1)(G-O)+G}\}$,  
where $G$ is the number of sentences per chunk and $O$ is sentence overlap. \\

Paragraph-Based and Paragraph-Group Chunking (PBC / PGC) &
Chunks align with paragraph boundaries or groups of adjacent paragraphs. &
Let $\{p_1,\dots,p_m\}$ denote paragraphs.  
$C_i = \{p_{(i-1)(G-O)+1},\dots,p_{(i-1)(G-O)+G}\}$,  
where $G$ is the number of paragraphs and $O$ is overlap. \\

Recursive Chunking (RC / RTF) &
Documents are recursively segmented using increasingly fine rules until a size constraint is met, with fallback token splitting if required. &
Given segment $x$, recursively apply  
$\text{split}(x)$ until $|C_i| \le L$,  
where $L$ is the maximum allowable chunk size (tokens or characters). \\

Parent--Child Chunking (PCC) &
A hierarchical structure is constructed in which parent chunks preserve broader context and child chunks serve as the atomic retrieval unit. &
Parent chunks $P_j$ satisfy $|P_j| \le L_p$.  
Each parent is subdivided into child chunks  
$C_{j,k} \subset P_j$, with $|C_{j,k}| \le L_c$,  
where $L_p > L_c$. \\

Dynamic Token Size Chunking (DFC / LAC) &
Chunk sizes vary dynamically within predefined bounds to balance granularity and context. &
Each chunk size $K_i$ is selected such that  
$K_i \in [K_{\min}, K_{\max}]$,  
where $K_{\min}$ and $K_{\max}$ are minimum and maximum token limits. \\

Content Density Adaptive Chunking (CDAC) &
Chunk sizes are adjusted inversely to local lexical density, producing smaller chunks in information-dense regions. &
Lexical density is defined as  
$\rho(x) = \frac{|V(x)|}{|x|}$,  
where $V(x)$ is the set of unique tokens in segment $x$.  
Chunk size scales as $K_i \propto \rho(x)^{-1}$. \\

Semantic Similarity Threshold Chunking (SSTC) &
Semantic boundaries are introduced when adjacent segments become dissimilar in embedding space. &
Let $f(\cdot)$ be an embedding function.  
$\mathrm{sim}(x_i,x_{i+1}) = \frac{f(x_i)\cdot f(x_{i+1})}{\|f(x_i)\|\|f(x_{i+1})\|}$.  
A boundary is inserted if $\mathrm{sim}(x_i,x_{i+1}) < \theta$,  
where $\theta$ is a similarity threshold. \\

Semantic Variance Adaptive Chunking (SVAC) &
Boundaries are triggered when semantic similarity drops significantly relative to recent context. &
A rolling mean similarity $\mu$ is maintained.  
A boundary is introduced if  
$\mathrm{sim}(x_i,x_{i+1}) < \mu - \delta$,  
where $\delta$ controls sensitivity to semantic change. \\

Topic-Based Chunking (TBC) &
Chunks correspond to contiguous regions sharing the same inferred topic. &
Sentences are embedded and clustered.  
A boundary is introduced when  
$\mathrm{cluster}(x_i) \neq \mathrm{cluster}(x_{i+1})$. \\

LLM Boundary Detection (LBDC / LSTC) &
Chunking is formulated as a learned boundary detection task using a language model. &
The model predicts boundary probability  
$p(b_i = 1 \mid x_i, x_{i+1})$.  
A boundary is inserted if $p(b_i=1) > \tau$,  
where $\tau$ is a decision threshold. \\

Late Chunking Strategies (LCTS / LCSI / LCPI) &
Chunk construction is deferred until retrieval or query time to preserve full-document context. &
Chunks are materialised as  
$C_i = \text{segment}(D \mid q)$,  
where $q$ is the retrieval query and segmentation depends on query relevance. \\

Hybrid Chunking Strategies &
Two-stage pipelines combining semantic or structural segmentation with strict size normalisation. &
$\mathcal{C} = \mathcal{S}(\mathcal{P}(D))$,  
where $\mathcal{P}$ produces coherent segments and  
$\mathcal{S}$ enforces token or character length constraints. \\

\bottomrule
\end{tabular}

\label{table2}
\end{table*}

%% file: table_3.tex
\begin{table*}[htbp]
\centering
\caption{Performance trade-offs for all 36 evaluated chunking strategies, sorted in descending order by chunking phase RAM. While all methods maintain an average query time of under 7 milliseconds, the preprocessing costs vary severely. LLM-based strategies (e.g., LSTC, LBDC) demand extreme memory overhead, and complex semantic clustering (e.g., TBC, HSSC) requires significantly higher processing time. Conversely, high-performing structural methods like Paragraph Group Chunking (PGC) remain highly resource efficient across all three metrics.}
\label{tab:memory_time_costs}
\begin{tabular}{@{}lrrr|lrrr@{}}
\toprule
\textbf{Strategy} & \textbf{RAM Cost} & \textbf{Chunking} & \textbf{Query Time} & \textbf{Strategy} & \textbf{RAM Cost} & \textbf{Chunking} & \textbf{Query Time} \\
 & \textbf{(MB)} & \textbf{Time (s)} & \textbf{(ms)} & & \textbf{(MB)} & \textbf{Time (s)} & \textbf{(ms)} \\
\midrule
\textbf{LSTC} & 5,672.16 & 10.02 & 3.84 & \textbf{LCTS} & 929.34 & 7.77 & 5.28 \\
\textbf{LBDC} & 5,417.98 & 9.27 & 3.62 & \textbf{LCSI} & 926.96 & 9.77 & 3.92 \\
\textbf{TBC} & 2,849.53 & 133.79 & 4.37 & \textbf{HPGC} & 925.70 & 8.52 & 4.27 \\
\textbf{HSSC} & 1,744.49 & 77.40 & 4.02 & \textbf{SBC} & 922.47 & 9.52 & 3.88 \\
\textbf{HSmFC} & 1,700.33 & 75.15 & 4.40 & \textbf{FC} & 915.20 & 7.77 & 4.02 \\
\textbf{SBDC} & 1,658.81 & 73.64 & 3.62 & \textbf{HSGC} & 914.89 & 9.02 & 3.86 \\
\textbf{SEBC} & 1,657.18 & 73.64 & 3.73 & \textbf{HPFC} & 891.13 & 6.77 & 4.05 \\
\textbf{HSVFC} & 1,601.20 & 74.39 & 4.68 & \textbf{HDFC} & 883.62 & 7.27 & 4.47 \\
\textbf{SSTC} & 1,594.77 & 75.90 & 3.99 & \textbf{DFC} & 883.49 & 7.27 & 5.49 \\
\textbf{SVAC} & 1,557.51 & 72.14 & 4.17 & \textbf{SGC} & 880.71 & 8.77 & 4.02 \\
\textbf{PCC} & 994.77 & 14.53 & 3.92 & \textbf{HRC} & 879.47 & 7.77 & 3.88 \\
\textbf{SWC} & 981.46 & 9.52 & 4.93 & \textbf{HCDC} & 874.56 & 19.54 & 4.65 \\
\textbf{HFCF} & 969.30 & 8.27 & 5.57 & \textbf{PGC} & 873.35 & 6.26 & 3.72 \\
\textbf{FCC} & 953.14 & 8.27 & 4.47 & \textbf{LCPI} & 859.46 & 6.01 & 3.75 \\
\textbf{HSnFC} & 942.21 & 10.02 & 4.00 & \textbf{PBC} & 858.21 & 6.01 & 5.45 \\
\textbf{RTF} & 932.91 & 19.79 & 3.89 & \textbf{RC} & 850.11 & 7.52 & 3.94 \\
\textbf{OFC} & 931.69 & 8.27 & 4.01 & \textbf{LAC} & 841.39 & 7.77 & 4.64 \\
\textbf{HOFC} & 930.66 & 8.52 & 4.03 & \textbf{CDAC} & 841.27 & 19.29 & 6.24 \\
\bottomrule
\end{tabular}
\label{table3}
\end{table*}